\documentclass{svproc}
\usepackage{url}

\usepackage{amsfonts}
\usepackage[margin=1.2in]{geometry}
\usepackage{graphicx}
\usepackage{multirow}
\usepackage{float}
\usepackage{lastpage}
\usepackage{fancyhdr}

\usepackage{colortbl}
\usepackage{color}
\usepackage{array, makecell}
\usepackage{boldline}
\usepackage[x11names, table]{xcolor}

\newcolumntype{?}[1]{!{\vrule width #1}}
\definecolor{colT1}{rgb}{0.95,0.88,0.2}
\definecolor{colT2}{rgb}{0.98,0.95,0.8}
\definecolor{colT3}{rgb}{0.0,0.0,0.0}
\definecolor{colT4}{rgb}{1.0,1.0,1.0}

\begin{document}
\mainmatter              
\title{ Methods of the Vehicle Re-identification}
\titlerunning{Vehicle re-identification}  
%
\author{Mohamed Nafzi \and Michael Brauckmann\inst{1} \and Tobias Glasmachers\inst{2}}
\authorrunning{Mohamed Nafzi et al.} 
%
\tocauthor{Ivar Ekeland, Roger Temam, Jeffrey Dean, David Grove,
Craig Chambers, Kim B. Bruce, and Elisa Bertino}
\institute{Facial \& Video Analytics\\
	IDEMIA Identity \& Security Germany AG\\
\email{mohamed.nafzi@idemia.com, michael.brauckmann@idemia.com},\\
\and
Institute for Neural Computation\\
Ruhr-University Bochum, Germany\\
\email{tobias.glasmachers@ini.rub.de}}

\maketitle              

\begin{abstract}
Most of researchers use the vehicle re-identification based on classification. This always requires an update with the new vehicle models in the market. In this paper, two types of vehicle re-identification will be presented. First, the standard method, which needs an image from the search vehicle. It produces a feature vector, which will be applied by the re-identification of the search vehicle. VRIC and VehicleID data set are suitable for training this module. It will be explained in detail how to improve the performance of this method using a trained network, which is designed for the classification. The second method takes as input a representative image of the search vehicle with similar make/model, released year and colour. It is very useful when an image from the search vehicle is not available. It produces as output a shape and a colour features. This could be used by the matching across a database to re-identify vehicles, which look similar to the search vehicle. To get a robust module for the re-identification, a fine-grained classification has been trained, which its class consists of four elements: the make of a vehicle refers to the vehicle’s manufacturer, e.g. Mercedes-Benz, the model of a vehicle refers to type of model within that manufacturer’s portfolio, e.g. C Class, the year refers to the iteration of the model, which may receive progressive alterations and upgrades by its manufacturer and the perspective of the vehicle. Thus, all four elements describe the vehicle at increasing degree of specificity. The aim of the vehicle shape classification is to classify the combination of these four elements. The colour classification has been separately trained. After the training, the classification layer will not be used. By both methods, even data of vehicles by some makes/models/released years/perspectives or by some colours are not available, it will be possible to re-identify each vehicle. The results of vehicle re-identification will be shown. Using a developed tool, the re-identification of vehicles on video images and on controlled data set using a search image will be demonstrated. The results of a proposed mix-mode, which is the combination of shape matching and colour classification, will be presented. This work was partially funded under the grant.
\end{abstract}

\section{Introduction}

The objective of the vehicle re-identification module is to recognize a vehicle within a large image
or video data set. Two different methods will be trained and tested.

\begin{itemize}

\item First, the standard vehicle re-identification. The known data set VRIC and VehicleID have been used separately for training and testing. VRIC data set contains 2811 vehicle-IDs with 54808 images and VehicleID contains 13164 vehicle-IDs with 113346 images for training. Also Multiple loss and a merged data set have been used to train on both data set. This, can increase the robustness of the module. Starting the training from a trained network, which has been trained on shape classification using about eight million images, can significantly improve the results. The results of the fusion will be also presented.

\item In the training of the second method, which requires just a representative image looks similar to the search vehicle in case its sample image is not available, a fine-grained vehicle classification has been used, which leads to feature representation with small intra-class variance. The modules have been trained using CNN-Networks. The combination of the shape and the colour feature vectors leads to a robust re-identification of vehicles.
\begin{itemize}
\item Training:
Typically, a fine-grained class consists of four elements: the make of a vehicle refers to the vehicle’s manufacturer, e.g. Mercedes-Benz, the model of a vehicle refers to type of model within that manufacturer’s portfolio, e.g. C Class, the year refers to the iteration of the model, which may receive progressive alterations and upgrades by its manufacturer and the perspective of the vehicle. Thus, all four elements describe the vehicle at increasing degree of specificity. The aim of the vehicle shape classification is to classify the combination of these four elements. We trained our vehicle shape network on 11906 classes using about eight million images. We trained the colour classification separately on 10 classes using about two million images.

\item Application:
In the application of our trained CNN-Network, the classification layer will not be used. Our module supports searches using an image sample or a representative image of the search vehicle, which is sent to the template creation component. The search engine performs the template matching across a video database using shape and colour features and returns the search results to the user. This method does not require the training of all vehicle classes. To get an alarm the make, the model, the released year, the perspective and the colour of the probe and of the gallery images should be similar.
\end{itemize}
\end{itemize}
\section{Related Works}
Some research has been performed on make/model classification to re-identify a search vehicle. Most of it operated on a small number of make/models because it is difficult to get a labeled data set panning all existing make/models. Manual annotation is almost impossible because one needs an expert for each make being able to recognize all its models and it is very tedious and time consuming process. \cite{paper14} developed a make/model classification based on feature representation for rigid structure recognition using 77 different classes. Two distances have been tested, the dot product and the euclidean distance. \cite{paper5} tested different methods by make/model classification of 86 different classes on images with side view. The best one was HoG-RBF-SVM. \cite{paper17} used 3D-boxes of the image with its rasterized low-resolution shape and information about the 3D vehicle orientation as CNN-input to classify 126 different make/models. The module of \cite{paper10} is based on 3D object representations using linear SVM classifiers and trained on 196 classes. In a real video scene all existing make/models could occur. Considering that we have worldwide more than 2000 models, make/model classification trained just on few classes will not succeed in practical applications. \cite{paper1} increase the number of the trained classes. His module is based on CNN and  trained on 59 different vehicle makes as well as on 818 different models. His solution seems to be closer for commercial use. Our developed module in our previous work \cite{paper18} was trained on 1447 different classes and could recognize 137 different vehicle makes as well as 1447 different models of the released year between 2016 till 2018. Other research has been operated on the known standard vehicle re-identification. Space-time contextual knowledge has been exploited for vehicle re-id subject to structured scenes. \cite{paper20} incorporated spatio-temporal path information of vehicles. This method improves the re-id performance on the VeRi-776 data set, it may not generalize
to complex scene structures when the number of visual spatio-temporal path proposals is very large with only weak contextual knowledge available to facilitate model decision.
\cite{paper21} considered 20 vehicle key points for learning and aligning local regions of a vehicle for re-identification. Clearly, this approach comes with extra cost of exhaustively labeling these key points in a large number of vehicle images, and the implicit assumption of having sufficient image resolution/details for computing these key points. \cite{paper22} worked on VehicleID data set, which includes multiple images of the same vehicle captured by different real world cameras in a city. This data set is challenging in term to separate between similar vehicles with few of differences but it is only constrained test scenarios due to the rather artificial assumption of having high quality images of constant resolution. This makes them limited for testing the true robustness of re-id matching algorithms in typically unconstrained wide-view traffic scene imaging conditions. \cite{paper19} introduced the Veric data set to address the limitation of other Vehicle re-identification Benchmarks, which provides conditions giving rise to changes in resolution, motion blur, weather, illumination, and occlusion. In this paper, we show two methods of the vehicle re-identification, which could re-identify vehicles even if their classes are not included in the training. First method is the standard vehicle re-identification, which requires a probe image of the search vehicle. This module has been trained using a merged data set of Veric and VehicleID. Its training has been started from the trained make/model network used in the second method, which is trained on classification using 11906 classes with about eight million images for the shape and using 10 classes with about two million images for the colour. It uses shape and colour feature vectors for the re-id. It works even if a probe image of the search vehicle is not available. A representative image with similar make, model, released year and colour of the search vehicle would be enough for the re-identification. It could be downloaded e.g. from the web. Experimental results show that the first method outperforms all state-of-the-art approaches on Veric and VehicleID data set. Here, the comparison has been done just to the best published results. The second method helps to improve the performance of the first method, and it gives a solution in case a probe image of the search vehicle is not available. Here, there are no defined data set we could use to compare the results to other research. Tests has been evaluated on an internally data set.

\section{ Network Architectures and Feature Extraction}
\begin{figure}[H]
	\centering
	\includegraphics[width=0.5\textwidth]{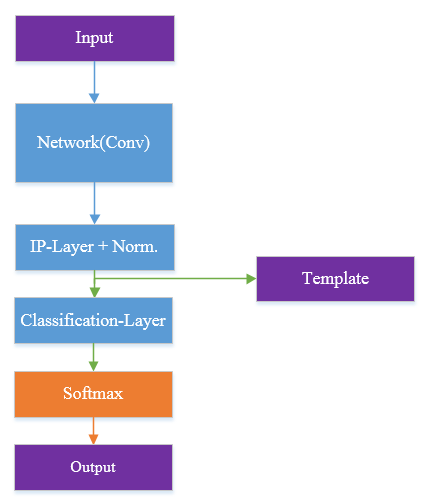}
	\caption{ Feature vector extraction. The network CNN1 for the vehicle re-identification based on shape and colour classification (method 2). Trained on 11906 classes for shape and on 10 classes for colour.}
	\label{net1}
\end{figure}

\begin{figure}[H]
	\centering
	\includegraphics[width=0.7\textwidth]{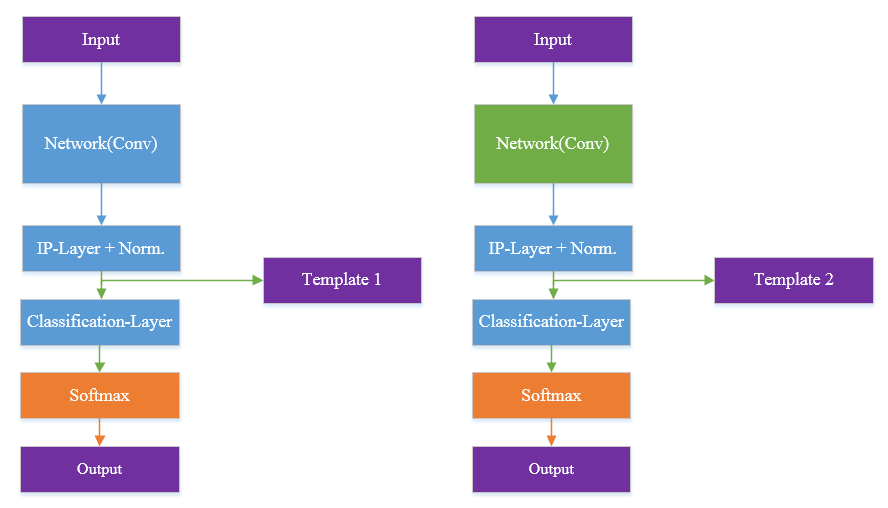}
	\caption{ Feature vector extraction. The network CNN2 for the standard vehicle re-identification (method 1). \newline Starting from a trained CNN (using trained CNN1 from the method 2).
	\newline Blue indicates trainable parameters. Green shows not trainable parameters.
	\newline Both CNNs are trained on a merged data set of Veric and VehicleID.
     \newline CNN2 is the fusion of CNN-nets and shows the best results.}
	\label{net2}
\end{figure}

Neural networks have been used in computer vision for a long time, but with the progress in hardware capabilities and growth of available training data over the last few years deep neural networks have become the most successful methods for many computer vision tasks. In some visual recognition tasks, even human-level accuracy can be surpassed. We used a CNN-networks based on ResNet architecture. Their coding time is 20ms
 (CPU 1 core, i7-4790, 3.6 GHz). In the figures \ref{net1}
 and \ref{net2}, we show our way to extract the feature vector, which will be used in the matching step by the vehicle re-identification. The figure \ref{net1} shows the trained CNN for the vehicle re-identification based on shape and colour classification (method 2), and the figure \ref{net2} shows the trained CNN for the the standard vehicle re-identification (trained on gray images / method 1). Here, we started from the trained CNN from the method 2, which has been trained on 11906 classes with about eight million images. This CNN-net is an expert to separate between vehicles with different makes, models or released years. by this way the training is focusing to separate between different vehicles with similar makes, models and released years but without forgetting to separate between vehicles with different makes, models or released years. Here, two CNN-nets have been trained. By the first training all parameters are trainable. By the second CNN-net the convolution block is not trainable. Here, the training tunes just the IP-Layer for the separation between the classes. The fusion shows the best results on Veric and VehicleID.

\section{Matching and fusion}
Our feature vectors (templates) are normalized to unit length. The matching as such is performed by calculating the dot product between two feature vectors
which i.e. the cosine of the angle between both vectors. Hereby by the method 2, the matching scores of the color and of
the shape feature vectors have different distributions. Fusion uses a weighted sum of the match scores of
both modalities. Optimal weights have been determined based on a predefined set of data. By method 1, the fusion score is the sum of the match scores, which have similar distributions.

\section{Mix-Mode}
According to our method 2 for vehicle re-identification based on shape and colour features, we need for a vehicle search a respective search
image of a certain make/model, released year and color. The make and model of the search image does not need to be
part of the make/models categories used during training.. In practice, we could have the case that we have
an image just with the same shape but not with the same color of the search vehicle, e.g. downloaded from
a manufacturer’s internet homepage. In this case, we could apply our developed Mixed-Mode, which is the
appropriate solution for this problem. In this mode, we combine the shape matching together with color
classification. We use the shape feature vector for matching. As results, we get all vehicles that have the
same shape as the searched vehicle however potentially with different colors. After that, we apply the color
classification to filter the results by the selected color. This mode is intended specifically to be used in investigational scenarios.

\section{Experiments}
\subsection{Experiments of the vehicle re-identification based on shape and colour features}
\begin{figure}[bth]
	\centering
	\begin{minipage}[c]{0.30\textwidth}
		\includegraphics[width=\linewidth]{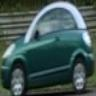}
	\end{minipage}
	\begin{minipage}[c]{0.30\textwidth}
		\includegraphics[width=\linewidth]{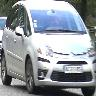}
	\end{minipage}
	\vspace*{2mm}
	\caption{The image on the left side shows a sample of a best-shot computed from the VICTORIA data set (“Cam2”). The image on the right side depicts a best-shot from “Cam4” respectively.}
	\label{s1}
\end{figure}

\begin{figure}[bth]
	\centering
	\begin{minipage}[c]{0.30\textwidth}
		\includegraphics[width=\linewidth]{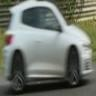}
	\end{minipage}
	\begin{minipage}[c]{0.30\textwidth}
		\includegraphics[width=\linewidth]{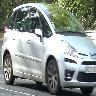}
	\end{minipage}
	\vspace*{2mm}
	\caption{The color “silver” is not included in our training of color classification.
		Right vehicle is labeled as gray but with sunlight looks close to white. It
		produces higher impostor scores with white vehicles like the vehicle on the left,
		this leads to a reduction of the verification rate as depicted by the black ROC
		curves in figures \ref{ri1} and \ref{ri2}}
	\label{image}
\end{figure}

\begin{figure}[H]
	\centering
	\includegraphics[width=0.72\textwidth]{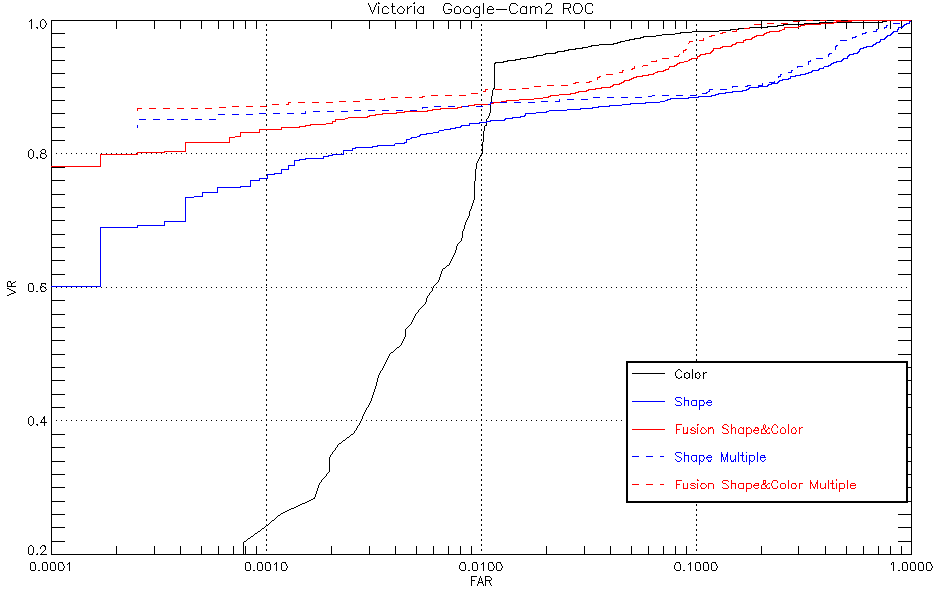}
	\caption{ This figure shows ROC-curves of shape, color, fusion of color and shape and using multiple probe images by shape. Computation was done matching of controlled single images from the internet against video data set Cam2 from the project Victoria. 
	\newline \textbf{Color}: matching using color template (black curve).
  	 \textbf{Shape}: matching using shape template (blue solid curve).
  	\newline \textbf{Fusion Shape\&Color}: Fusion of shape and color matching scores (red solid curve).
  	\newline \textbf{Shape Multiple}: matching using shape template and using multiple probe images (blue dashed curve).
  	\newline \textbf{Fusion Shape\&Color Multiple}: Fusion of shape using multiple probe images and color matching scores (dashed solid curve).
    \newline \textbf{FAR}: False Acceptance Rate. \quad
     \textbf{VR}: Verification Rate.}
	\label{ri1}
\end{figure}

\begin{figure}[H]
	\centering
	\includegraphics[width=0.72\textwidth]{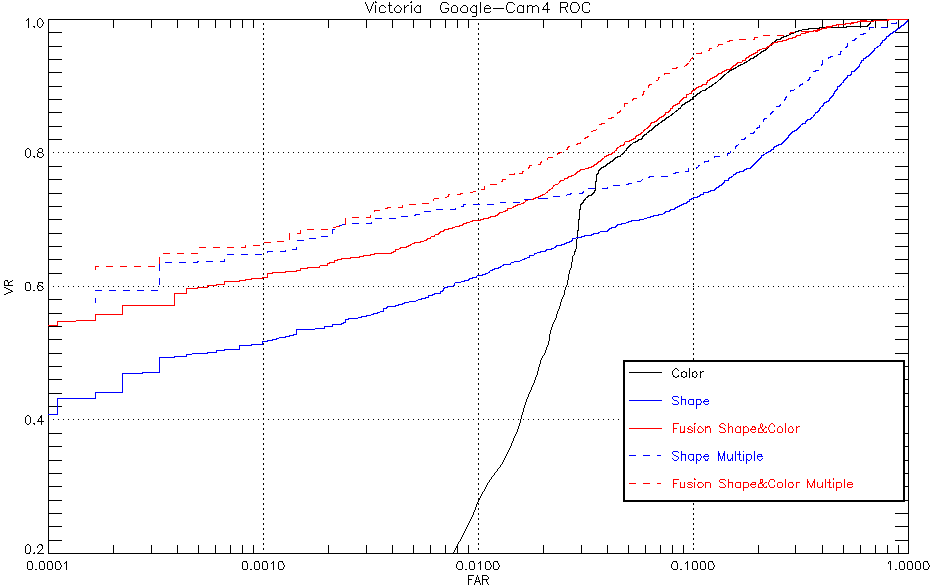}
	\caption{ This figure shows ROC-curves of shape, color, fusion of color and shape and using multiple probe images by shape. Computation was done matching of controlled single images from the internet against video data set Cam4 from the project Victoria. 
		\newline \textbf{Color}: matching using color template (black curve).
		\textbf{Shape}: matching using shape template (blue solid curve).
		\newline \textbf{Fusion Shape\&Color}: Fusion of shape and color matching scores (red solid curve).
		\newline \textbf{Shape Multiple}: matching using shape template and using multiple probe images (blue dashed curve).
		\newline \textbf{Fusion Shape\&Color Multiple}: Fusion of shape using multiple probe images and color matching scores (dashed solid curve).
		\newline \textbf{FAR}: False Acceptance Rate. \quad
		\textbf{VR}: Verification Rate.}
	\label{ri2}
\end{figure}

In total, 406 best-shots and 85.130 detections were computed from Cam2, and 621 best-shots with
199.963 detections from Cam4. Additionally, 33 controlled images were acquired from the web (“Google”)
for subsequent experiments. Based on these VICTORIA data sets, we performed a number of tests using the shape feature, the colour feature and the fusion of both. multiple probe images by shape matching have been also tested. Here, we have a set of images of the search vehicle with different views. By matching across a gallery image, we get a set of scores. Their maximum is the finale match score. This reduces the dependency of the perspective by matching. Tests have been evaluated on video data across still images. The figure \ref{s1} shows sample images from the video data set Cam2 and Cam4. Results are shown in the figures \ref{ri1} and \ref{ri2}. Here as shown, we got some high impostor scores by matching of color templates, leading to a fall of the ROC curves. The reason for this is that the
color “silver” is currently not included in the classes used for the training, thus we labelled it as “grey”. Due
to the sun-light conditions however, the silver color was mapped onto “white”. The figure \ref{image} shows
two sample images illustrating this effect.

\subsection{Experiments of the standard vehicle re-identification}
\textbf{Data sets}. For evaluation, we utilised two most popular vehicle re-identification benchmarks. The VehicleID data set
\cite{paper22} provides a training set with 113,346 images from 13,164 IDs and a test set
with 17,377 probe images and 2,400 gallery images from 2,400 identities. It adopts the single-shot re-id setting,
with only one true matching for each probe. The VRIC data set \cite{paper19} has 54,808 images from 2,811 IDs in training set.
The probe and the gallery of the testing data set contain 2,811 images with 2,811 vehicle IDs. The data split statistics are summarised in table \ref{tb1}.
\newline \textbf{Evaluation}. Table \ref{tb2} compares our method1 (CNN2) explained in sections before with state-of-the-art methods on two benchmarks. Our method outperforms all other competitors with large margins. It surpasses the best competitor in Rank-1 rate by 8.53\% (this means 16.0\% error reduction) and in Rank-5 by 9.55\% on VRIC, and in Rank-1 rate by 2.8\% (this means 7.6\% error reduction) and in Rank-5 by 4.2\% on VehicleID.

\begin{table}[H]
	\caption{Data split of standard vehicle re-identification data sets evaluated in our experiments.} 
	\begin{tabular}{ ?{0.7pt}c?{0.7pt}c?{0.7pt}c?{0.7pt}c?{0.7pt} }
		\arrayrulecolor{colT3}\hline
		\hlineB{1.5}\cellcolor{colT4} Dataset &\cellcolor{colT4} Training IDs / Images &\cellcolor{colT4} Probe IDs / Images & \cellcolor{colT4} Gallery IDs / Images \\ 
		\hline
		\hlineB{1.6}\cellcolor{colT4} VehicleID \cite{paper22} & \cellcolor{colT4} 13,164 / 113,346 & \cellcolor{colT4}  2,400 / 17,377 & \cellcolor{colT4}  2,400 / 2,400 \\
		\hline			
		\hlineB{1.6}\cellcolor{colT4} VRIC \cite{paper19}& \cellcolor{colT4} 2,811 / 54,808 & \cellcolor{colT4} 2,811 / 2,811 & \cellcolor{colT4} 2,811 / 2,811 \\  
		\hlineB{1.4}\hline
		
	\end{tabular}
	\label{tb1}
\end{table}

\begin{table}[H]
	\caption{Comparative of standard vehicle re-identification results on two benchmarking data sets.} 
	\begin{tabular}{ ?{0.7pt}c?{0.7pt}c?{0.7pt}c?{0.7pt}c?{0.7pt} }
		\arrayrulecolor{colT3}\hline
		\hlineB{1.5}\cellcolor{colT4} Method &\cellcolor{colT4} VehicleID \cite{paper22} &\cellcolor{colT4} VRIC \cite{paper19}  \\ 
		\hline
		\hlineB{1.6}\cellcolor{colT4}  & \cellcolor{colT4} Rank-1 \quad   Rank-5 & \cellcolor{colT4} Rank-1 \quad   Rank-5 \\
		\hline
		\hlineB{1.6}\cellcolor{colT4} OIFE(Single Branch)\cite{paper21} & \cellcolor{colT4} 32.86 \quad   52.75 & \cellcolor{colT4} 24.62 \quad   50.98 \\
		\hline					
		\hlineB{1.6}\cellcolor{colT4} Siamese-Visual\cite{paper20}& \cellcolor{colT4} 36.83 \quad   57.97 & \cellcolor{colT4} 30.55 \quad   57.30  \\  
		\hline			
		\hlineB{1.6}\cellcolor{colT4} MSVF\cite{paper19}& \cellcolor{colT4} 63.02 \quad   73.05 & \cellcolor{colT4} 46.61 \quad   65.58  \\  
		\hline			
		\hlineB{1.6}\cellcolor{colT4} our method 1 (CNN2)& \cellcolor{colT4} \color{red} 65.82 \quad   77.25 & \cellcolor{colT4} \color{red} 55.14 \quad   75.13  \\  
		\hlineB{1.4}\hline
		
	\end{tabular}
	\label{tb2}
\end{table}

\section{Manual testing using our Vehicle re-identification tool}

\begin{figure}[H]
	\centering
	\includegraphics[width=0.72\textwidth]{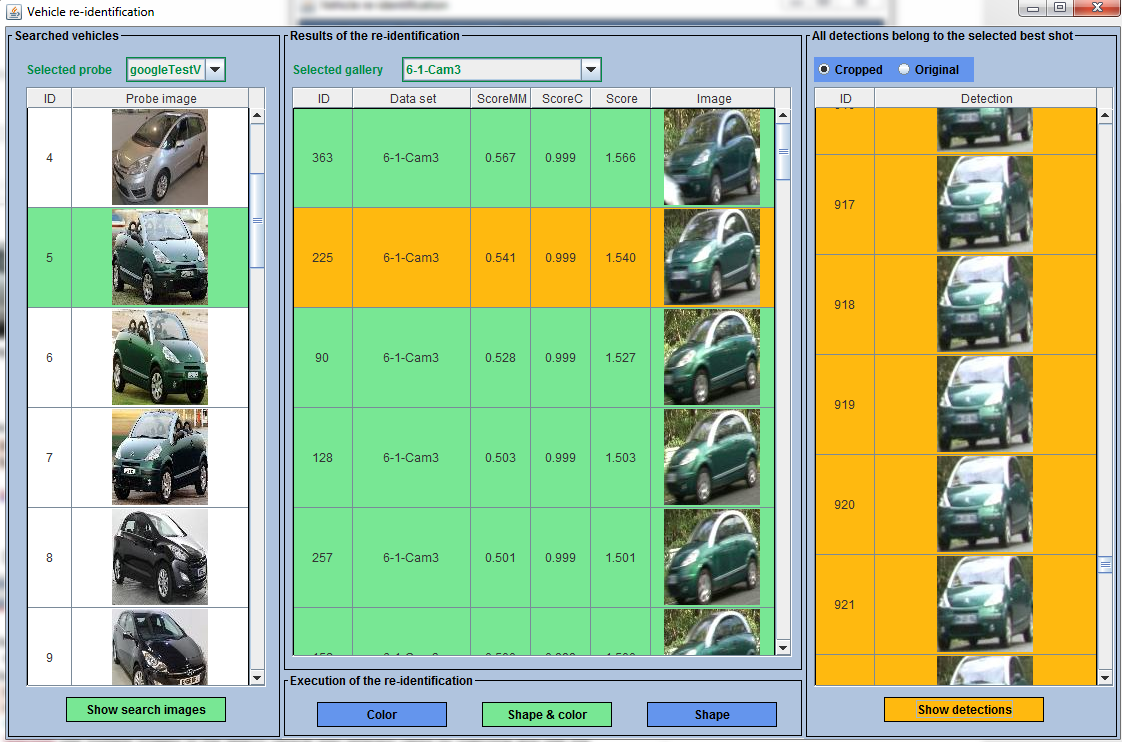}
	\caption{ This figure shows the vehicle re-identification based on shape and color features.}
	\label{tool1}
\end{figure}

\begin{figure}[H]
	\centering
	\includegraphics[width=0.72\textwidth]{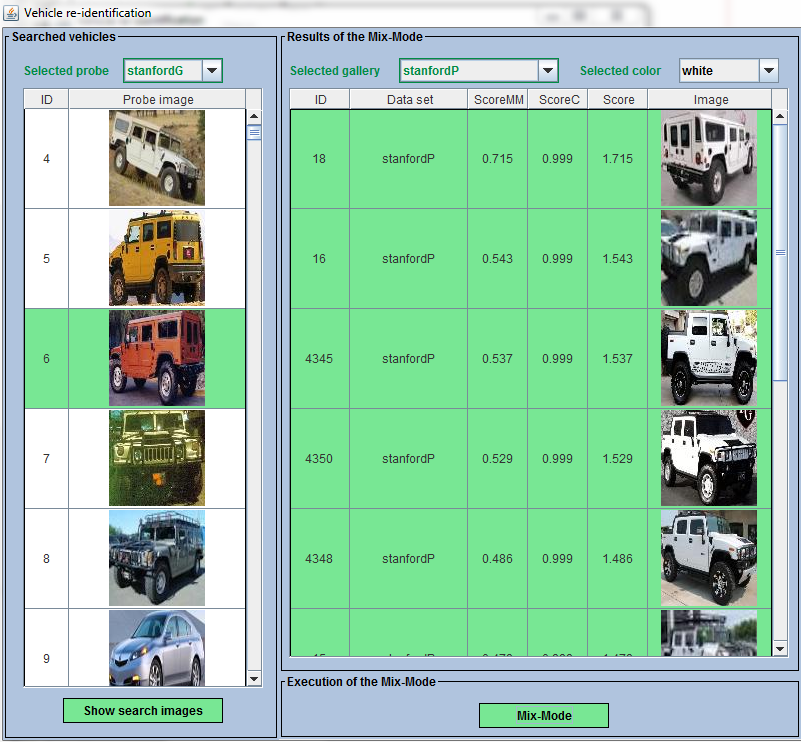}
	\caption{ This figure shows the vehicle re-identification using the Mix-Mode based on shape feature and color classification.}
	\label{tool2}
\end{figure}

Besides the statistical experiments from the section before, we performed manual tests on the second method trained on shape and colour features with the vehicle re-identification tool. We tested also the mix-mode, which has been defined in this research. The figure \ref{tool1} shows exemplary the search for a green “Ford Ka”. The left side of the figure depicts the selected search image, the middle part shows the best-shots of the matches against the VICTORIA data (Cam3 video sequence), and the right side presents all detections belonging to the selected best-shot. 
The subsequent Figure \ref{tool2} shows an example for the Mixed Mode. In this scenario, the user searches for a
‘white’ Hummer 2. In case that a sample image of that Hummer 2 is available, however with a different color, here ‘orange’, he nevertheless can apply the search
that provides all occurrences of that Hummer 2 however with any color. In a follow-up step,
color classification is applied to filter those result images with the searched color, here ‘white’. 

\section{Conclusion and Future Work}
\begin{itemize}
 \item Both vehicle re-identification methods work on classes even if they are not included in the training. They have not immediately to be updated with new released models.
 \item The perspectives of the probe and of the gallery samples by mates should be similar to get an alarm. Using multiple probe images with different views make the re-identification 
 independently of the perspective.
 \item Vehicle re-identification based on shape and colour classification works even if an image of the search vehicle is not available. A representative image is sufficient. It re-identifies all vehicles with similar makes, models, released years and colours.
 \item  An image of the search vehicle is required for the standard re-identification, which could re-identify exactly the same vehicle.
 \item  The training of the Vehicle re-identification based on shape classification helps the training of the standard re-identification because the size of the training data of the first training is much larger than the the second training. Its results beats the best published methods as shown in the table \ref{tb2}.
 \item We are working on the classification of the perspective of the vehicle based on image or template.
 \item We plan to augment training data for the standard vehicle re-identification.
 \item We are working on different methods to improve the vehicle shape classification.
\end{itemize}

\section{Acknowledgment}
\begin{itemize}
	\item Victoria: funded by the European Commission (H2020), Grant Agreement number 740754 and is for Video analysis for Investigation of Criminal and Terrorist Activities.
	\item Florida: funded by the German Ministry of Education and Research (BMBF).
\end{itemize}

%
%

\end{document}